\documentclass[conference]{IEEEtran}
\IEEEoverridecommandlockouts

\usepackage{cite}
\usepackage{amsmath,amssymb,amsfonts}
\usepackage{graphicx}
\usepackage{float}
\usepackage{placeins}
\usepackage{flafter}

\usepackage{capt-of}
\usepackage{textcomp}
\usepackage{xcolor}
\usepackage{booktabs}
\usepackage{array}
\usepackage{longtable}
\usepackage{tabularx}
\usepackage{xurl}
\usepackage{makecell}
\usepackage{hyperref}
\hypersetup{
    colorlinks=true,
    linkcolor=blue,
    citecolor=blue,
    urlcolor=blue,
    breaklinks=true,
    pdftitle={A Physics-Consistent Benchmark for Contact-Rich Human--Robot Interaction in Assistive Care},
}
\usepackage{ragged2e}

\usepackage[table]{xcolor}

\definecolor{cellgreen}{RGB}{198,239,206}
\definecolor{cellyellow}{RGB}{255,235,156}
\definecolor{cellred}{RGB}{255,199,206}

\usepackage{pifont}
\usepackage{xcolor}
\usepackage{amssymb}

\definecolor{deepgreen}{RGB}{0,100,0}
\definecolor{deepred}{RGB}{180,0,0}
\definecolor{amber}{RGB}{210,150,0}

\newcommand{\cmark}{\textcolor{deepgreen}{\ding{51}}}
\newcommand{\xmark}{\textcolor{deepred}{\ding{55}}}
\newcommand{\pmark}{\textcolor{amber}{$\blacksquare$}}

\usepackage{url}

\newcommand{\code}[1]{\texttt{#1}}

\newcommand{\repo}[1]{{\small\path{#1}}}

\begin{document}

\title{A Physics-Consistent Benchmark for Contact-Rich Human--Robot Interaction in Assistive Care}

\ifdefined\ANONYMOUS
\author{}
\hypersetup{pdfauthor={}}
\else
\author{
\IEEEauthorblockN{Chengxiao He\textsuperscript{1,\textdagger}, Shenghai Yuan\textsuperscript{2,\textdagger}, Liuqun Fan\textsuperscript{1,*}, and Shenzhe Zhu\textsuperscript{3}}
\IEEEauthorblockA{\textsuperscript{1}\textit{School of Mechanical and Robotics Engineering, Tongji University}, Shanghai, China\\
\textsuperscript{2}\textit{Centre for Advanced Robotics Technology Innovation, Nanyang Technological University}, Singapore\\
\textsuperscript{3}\textit{College of Electronic and Information Engineering, Tongji University}, Shanghai, China\\
\{2130238,lqfan,2531819\}@tongji.edu.cn, shyuan@ntu.edu.sg}
\thanks{\textsuperscript{\textdagger}Equal contribution.}
\thanks{*Corresponding author: Liuqun Fan.}
}
\fi

\maketitle


\begin{abstract}
Conventional task-level evaluation asks whether a robot policy completes a specified action, but can miss failures that emerge only during physical human contact. This limitation is critical in contact-rich assistive tasks, where meaningful evaluation requires a physically responsive human, interaction-quality assessment beyond task success, and a leak-free observation protocol. We introduce a physics-consistent benchmark for contact-rich human--robot interaction, instantiated in robot-assisted bathing. The benchmark combines a deformable, passively responding human, physics-aware scores alongside task-level success, and a frozen vision-only / scorer-only evaluation protocol. To establish physical validity, region-wise simulated responses are calibrated against force--indentation measurements from Franka impedance pushes on a medical-care manikin. Under a frozen T1--T7 protocol with 140 runs per method, an LLM-augmented state machine (State Machine) achieves $72.9\%$ task success but drops to $56.4\%$ after correct-region and force-safety screening; VoxPoser produces lighter and more stable contact but completes only $27.9\%$ of trials; and zero-shot $\pi_{0.5}$ achieves $0.7\%$ task success with no correct-region or safety-gated successes. These results show that task completion alone does not imply physically valid contact and motivate physics-aware screening before deployment of contact-rich assistive robot policies. Code and benchmark assets are available at \url{https://anonymous.4open.science/r/Physics-Consistent-Benchmark_4_HRC-8DBF/}.
\end{abstract}

\section{Introduction}
\label{sec:introduction}

Contact-rich human--robot interaction poses an evaluation challenge that conventional task-level metrics do not capture. This challenge is particularly evident in assistive-care tasks such as robot-assisted bathing, where physical contact is intrinsic to task execution. Unlike tabletop manipulation, where the environment can often be treated as rigid and success can be decided geometrically, physical interaction with a human body involves deformation, passive joint motion, contact force, and joint-limit constraints. A policy may reach the instructed body region while exerting excessive force, approaching a joint limit, or failing to maintain stable contact. Task completion therefore does not imply physically valid contact.

Despite progress in assistive robotics and human--robot interaction, existing benchmarks remain limited in three respects. The physical response of the human is often simplified as rigid or motion-replayed; for example, safety-aware HRC benchmarks such as Human-Robot Gym incorporate learning and online constraints but evaluate interaction with a motion-driven human~\cite{thumm2024humanrobotgym}, preventing contact-induced deformation, force transmission, and passive joint motion from emerging. Meanwhile, care systems such as Manip4Care and VTTB, together with language-grounding and foundation-model pipelines including VoxPoser and $\pi_{0.5}$, primarily evaluate whether a task is completed through success, trajectory, or collision measures rather than how physical contact is executed~\cite{koh2025manip4care,gu2024vttb,openx2024rtx,zhang2025irefvla,huang2023voxposer,huang2024rekep,black2025pi05}. Finally, simulation-based protocols may expose privileged states such as ground-truth joint angles, soft-anchor states, or future contact labels to the tested policy, confounding policy capability with simulator access and breaking the intended sim-to-real observation contract.

These limitations raise three \textbf{challenges}: how to model a human physically responding to robot contact, how to evaluate interaction quality beyond task success, and how to separate policy-visible observations from scorer-only physical states.

\begin{figure*}[!t]
\centering
\includegraphics[width=\textwidth]{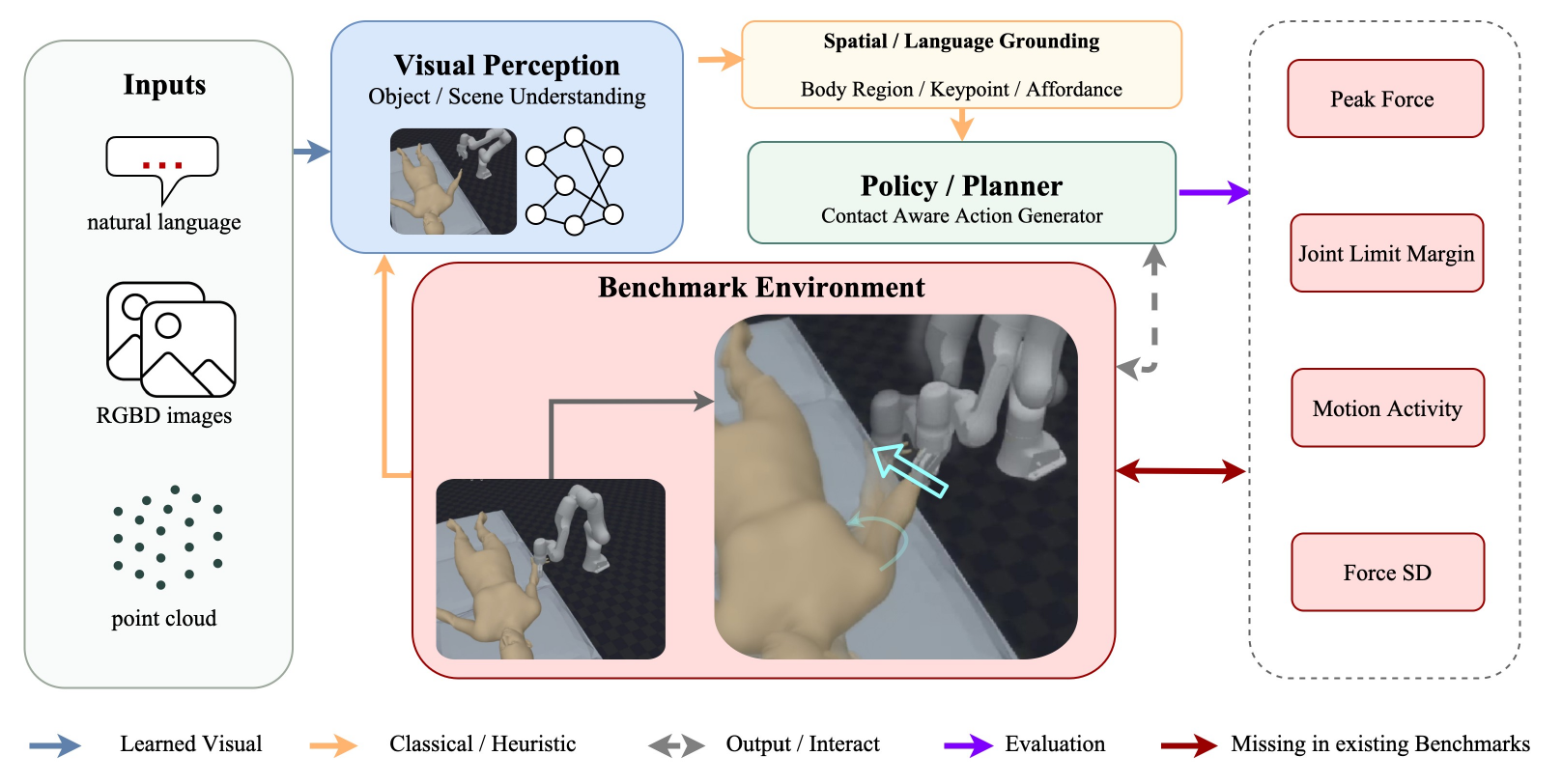}
\caption{Overview of the benchmark architecture and its contact-responsive human-in-the-loop evaluation. Language and RGB-D observations are processed by visual perception and spatial/language grounding before a policy/planner generates contact-aware robot actions. The benchmark environment contains an articulated human with deformable soft tissue and passive response; the two rendered interaction states illustrate contact-induced human motion. The resulting physical interaction is evaluated through C1 peak force, C2 joint-limit margin, C4 motion activity, and C5 force stability. Red links denote the physical-response and scoring pathway missing from conventional task-level benchmarks.}
\label{fig:benchmark_architecture}
\end{figure*}

To address these challenges, we introduce a physics-consistent benchmark for evaluating robot policies under contact-rich human--robot interaction in assistive care, instantiated and evaluated in robotic bathing. Figure~\ref{fig:benchmark_architecture} summarizes the resulting human-in-the-loop pipeline and illustrates passive human motion under robot contact. The benchmark combines a deformable, passively responding human, physics-aware scores with task-level success as a baseline, and a frozen vision-only / scorer-only protocol. To establish physical validity of the environment, we calibrate simulated region-wise responses against force--indentation measurements on a medical-care manikin. This correspondence grounds the benchmark; it is not offered as a standalone real-to-sim method. We do not introduce a new deformable solver: the physics stack is an implementation abstraction~\cite{macklin2016xpbd,chen2024vbd,liu2024softmac,liu2013skeletonsoftbody}, and fidelity is judged by measured contact agreement rather than by solver novelty.

The benchmark's finding is that task-level scores can hide contact failure. Post-hoc physical screening can distinguish nominal task success from deployment-usable success: State Machine drops from $102/140$ to $79/140$ under a safety gate, while $\pi_{0.5}$ retains $0/140$ gated success after $1/140$ original success. Our contributions are:
\begin{enumerate}
\item A physics-consistent benchmark for contact-rich human--robot interaction in assistive care, instantiated in robotic bathing as T1--T7 Touch / Scrub / Push.
\item A benchmark construction with three necessary parts: a deformable, passively responding environment; manikin-calibrated contact response; and a frozen vision-only / scorer-only protocol.
\item A diagnostic evaluation of State Machine, VoxPoser, and a zero-shot $\pi_{0.5}$ transfer probe revealing method-specific failure modes in force, joint-limit margin, contact stability, and region grounding, and showing that task completion does not imply physically valid contact.
\end{enumerate}
The benchmark implementation and evaluation configurations are available at \url{URL}.

Section~2 reviews the evaluation gap. Section~3 specifies the benchmark design. Section~4 reports environment validity and then the hidden failures. Section~5 states the finding.

\section{Related Work}
\label{sec:related_work}

\subsection{Care Systems and Policy Evaluation}

Early care systems such as I-SUPPORT explored pneumatic soft manipulators for elderly bathing~\cite{manti2016soft_assistive,manti2017towards_soft_manipulator}. Manip4Care addresses limb grasping and repositioning for bed bathing~\cite{koh2025manip4care}; VTTB studies visuo-tactile bed-bathing imitation on medical manikins~\cite{gu2024vttb}. These works show that robots can complete local care subtasks. They demonstrate task completion, not a reusable protocol that scores physical contact under a frozen observation contract. In parallel, language grounding and foundation VLA training have advanced rapidly~\cite{zhang2025irefvla,openx2024rtx}. VoxPoser and ReKep map language to spatial value maps or keypoint constraints~\cite{huang2023voxposer,huang2024rekep}; end-to-end models such as $\pi_0$, $\pi_{0.5}$, and OpenVLA improve continuous control and open-set generalization~\cite{black2024pi0,black2025pi05}. FAM-HRI further explores multimodal interaction for limited-mobility users~\cite{lai2025famhri}. Strong scores on general manipulation benchmarks, however, do not imply safe sustained contact with a deformable human body.

\subsection{Why Task-Level Benchmarks Miss Contact}

Human-Robot Gym provides a safety-aware HRC RL benchmark, but its human model is primarily motion-capture driven~\cite{thumm2024humanrobotgym}. Soft-body methods including XPBD, VBD, SoftMAC, skeleton--soft coupling, and industrial engines such as Newton improve physical realism~\cite{macklin2016xpbd,chen2024vbd,liu2024softmac,liu2013skeletonsoftbody,nvidia2026newton}. For this paper, such engines are \emph{infrastructure}: the claim is a physics-consistent benchmark with a calibrated human response. Accordingly, we judge the environment by real-to-sim agreement rather than by timestep or convergence contests against general-purpose engines.

On the scoring side, ISO~15066 specifies collaborative force limits by body region~\cite{iso15066}. Those limits are fixed rather than context-dependent, and the standard does not provide a reusable apparatus for scoring contact quality under a frozen observation contract. Behrens et al.\ report region-wise biomechanical pain-onset forces~\cite{behrens2021painlimits}; PrioriTouch discusses conservative scaling of those limits for contact preference~\cite{madan2025prioritouch}. Reviews of mental workload in HRI identify robot motion, rhythm, and predictability as workload-relevant factors~\cite{carissoli2024mentalworkload}. Language-conditioned manipulation suites such as LIBERO and dataset-scale policy evaluation such as Open X-Embodiment complement this literature, but they are not designed around passive human joints or calibrated contact force~\cite{liu2023libero,openx2024rtx}.

Existing systems provide important but largely separate ingredients, including care-task execution, deformable simulation, physical safety criteria, and policy evaluation. However, they do not jointly provide a deformable and passively responding human, calibrated contact response, leak-free policy observations, and physics-aware scoring within a unified evaluation protocol. Table~\ref{tab:capability} positions our benchmark with respect to these capabilities in the evaluated setting. Rather than introducing another task or simulation engine, our benchmark integrates these elements to evaluate whether successful contact-rich interaction is also physically valid.

\begin{table}[!t]
\centering
\caption{Comparison of benchmark-level capabilities relevant to our evaluation setting.}
\label{tab:capability}
\footnotesize
\setlength{\tabcolsep}{3pt}
\begin{tabularx}{\columnwidth}{
>{\raggedright\arraybackslash}p{0.30\columnwidth}
*{5}{>{\centering\arraybackslash}X}}
\toprule
System & DHS & PJR & FC & LFO & PAS \\
\midrule
Human-Robot Gym~\cite{thumm2024humanrobotgym}
& \xmark & \xmark & \pmark & \pmark & \pmark \\

LIBERO~\cite{liu2023libero}
& \xmark & \xmark & \xmark & \pmark & \xmark \\

Open X-Embodiment~\cite{openx2024rtx}
& \xmark & \xmark & \xmark & \pmark & \xmark \\

This benchmark
& \cmark & \cmark & \cmark & \cmark & \cmark \\
\bottomrule
\end{tabularx}

\vspace{2pt}
\raggedright
\scriptsize
DHS: deformable-human simulation;
PJR: passive-joint response;
FC: force calibration;
LFO: leak-free observation;
PAS: physics-aware scoring.
\cmark: supported;
\pmark: partially supported;
\xmark: not supported.
Open X-Embodiment is dataset-scale policy evaluation rather than a protocol benchmark.
\end{table}
\section{Benchmark Design}
\label{sec:benchmark}

The benchmark is specified as environment, tasks, then evaluation protocol. Detailed coupling derivations are provided in the supplementary material. We define physics consistency at the benchmark level rather than at the solver level. A benchmark is physics-consistent when (i)~its simulated human response exhibits measurable correspondence with real interaction within the evaluated contact regime, (ii)~contact-induced physical states---force, deformation, and passive joint motion---are dynamically generated rather than prescribed, and (iii)~these states are evaluated without exposing privileged simulator information to the tested policy. The resulting properties are physical realism, physical validity, evaluation validity, and diagnostic capability.

\subsection{Benchmark Environment}
\label{sec:human_modeling}

The benchmark requires a human that responds physically to contact rather than a kinematic target, together with an observation interface that evaluated policies can consume without simulator-state leakage.

\paragraph{Representation.}
\label{sec:human_representation}
The simulated care recipient combines an articulated rigid skeleton with region-wise deformable soft-tissue shells. Body regions follow the kinematic joint tree: load-bearing areas (forearm, upper arm, abdomen, and related torso/limb patches) use flexible surface shells anchored to associated bone segments, while low-compliance parts (e.g., fingers and head) may be treated as rigid. Passive joint behavior is always enabled so that contact forces can induce posture change without an active human controller. Local deformation, force transmission, and passive response are environment requirements.

\paragraph{Physical interaction.}
\label{sec:interaction_model}
We use an existing physics simulation framework as an implementation abstraction and construct a benchmark environment whose physical response can be calibrated against real human--robot contact. The following equations summarize shell--bone load transfer in our stack.
Contact on a soft region deforms the shell, transmits force to the skeleton through regional anchors, and can move passive joints. With skeleton coordinates $q$, stacked shell coordinates $x$, region-anchor Jacobians $J_{c,r}$, and stacked anchor reaction $f_c$,
\begin{equation}
M_r(q)\ddot{q} + C_r(q, \dot{q})\dot{q} + g_r(q)
 = \tau_{act} + \sum_{r=1}^{R} J_{c,r}(q)^T f_c^{(r)},
\label{eq:rigid_balance}
\end{equation}
\begin{equation}
M_f \ddot{x} + f_{int}(x, \dot{x})
 = f_{ext} - S^T f_c,
\label{eq:soft_balance}
\end{equation}
\begin{equation}
f_c = K_c \phi + D_c \dot{\phi},
\label{eq:coupling_force}
\end{equation}
where $S$ selects shell-side anchors, $f_c=\mathrm{col}(f_c^{(1)},\ldots,f_c^{(R)})$, and $\phi$ is the stacked anchor residual between shell and bone targets. The supplementary material provides symbol tables and solver variants. For evaluation, the consequence is behavioral: policies face deformation, tissue-mediated force, and passive joint motion---not a rigid mannequin.

\paragraph{Observation interface.}
\label{sec:observation_interface}
To keep evaluation leakage-free, the platform separates physics state from planner-visible perception. Rigid bones and flexible shells participate in mechanics only; an SMPL-X~\cite{pavlakos2019smplx} visual skin is driven unidirectionally from $q$ for rendering. A virtual camera and robot TCP pose feed a snapshot query interface that exposes:
\begin{enumerate}
\item RGB and depth images;
\item 3D point clouds;
\item zero-shot semantic segmentation masks for human, robot, and scene entities~\cite{ren2024grounded};
\item skeletal landmarks from vision-based keypoint estimators~\cite{lugaresi2019mediapipe}.
\end{enumerate}
Ground-truth joint angles, anchor states, and scorer-only contact metadata remain closed to evaluated policies. All semantic and spatial information used by a subject must be inferred from the vision-facing modalities above, which preserves a fair sim-to-real evaluation contract.

\subsection{Benchmark Tasks and Evaluation Dimensions}
\label{sec:tasks}

The suite emphasizes continuous physical interaction with deformable body regions rather than isolated rigid-object manipulation or language-ambiguity puzzles. We evaluate seven micro-tasks on three soft-relevant body regions. Unless noted, each trial starts from a supine care-recipient pose on a nursing bed with randomized comfort-arm orientation in a fixed angular range. Scrub tasks denote sustained sliding cleaning contact in the bathing sense (wipe family), not antipodal grasping.

\begin{table}[t]
\centering
\caption{Benchmark task set. Push on the upper arm is omitted because the short moment arm yields limited displacement diversity.}
\label{tab:task_set}
\footnotesize
\begin{tabularx}{\columnwidth}{>{\raggedright\arraybackslash}p{0.10\columnwidth} >{\raggedright\arraybackslash}p{0.28\columnwidth} >{\raggedright\arraybackslash}X}
\toprule
ID & Task & Purpose / primary cues \\
\midrule
T1 & Touch forearm & Contact establishment; localization; force overshoot \\
T2 & Push forearm & Deformation and displacement under external force \\
T3 & Scrub forearm & Sustained sliding cleaning contact; force stability; contact consistency \\
T4 & Touch upper arm & Region-dependent contact response \\
T5 & Scrub upper arm & Proximal soft-region scrubbing contact \\
T6 & Touch abdomen & Clear soft deformation; soft-contact necessity \\
T7 & Scrub abdomen & Soft-region scrubbing; pressure / force distribution \\
\bottomrule
\end{tabularx}
\end{table}

Natural-language prompts name the region and intent (e.g., ``gently touch the right forearm,'' ``push the forearm,'' ``gently scrub the abdomen''). Tasks that mainly probe rigid skeletal landmarks without soft-tissue involvement, language-ambiguity suites, and arm-reposition skills that reduce to conventional manipulation are excluded from this scoped task set.

\label{sec:metrics}
Scorer-only ground truth yields a task-level baseline and physics-aware scores (Table~\ref{tab:metric_defs}). These dimensions are particularly relevant to assistive-care interaction; they are defined to describe physical contact, not as clinically validated care scores.

Peak force is compared with region-wise biomechanical pain-onset limits $F_{\mathrm{pain}}(b)$ from Behrens et al.\ (Table~8, pinching force): $100$\,N on forearm muscle, $100$\,N on deltoid muscle, and $60$\,N on abdominal muscle~\cite{behrens2021painlimits}. All three task regions use muscle / soft-tissue entries, matching the soft-relevant sites of Section~\ref{sec:tasks}. We adopt a single conservative discount $\gamma'=0.35$ applied uniformly across regions, so $F_{\max}(b)=\gamma'F_{\mathrm{pain}}(b)$. The resulting safety gates are $35$\,N / $35$\,N / $21$\,N. In this work $F_{\mathrm{pain}}$ and $F_{\max}$ are two presentations of the same published limit---the literature pain threshold and a more conservative uniform deployment gate---not two independently measured quantities.

\begin{table}[t]
\centering
\caption{Evaluation dimensions. C3 is the task-level baseline; C1, C2, C4, and C5 are physics-aware. Symbols match the scorer implementation; all quantities are invisible to evaluated policies.}
\label{tab:metric_defs}
\footnotesize
\begin{tabularx}{\columnwidth}{>{\raggedright\arraybackslash}p{0.16\columnwidth} >{\raggedright\arraybackslash}p{0.18\columnwidth} >{\raggedright\arraybackslash}X}
\toprule
Layer & Metric & Definition \\
\midrule
Physics-aware & C1 Force & Peak $F_{peak}=\max_t\|F_c(t)\|$ during contact; compared with Behrens Table~8 $F_{\mathrm{pain}}(b)$ and the uniform $\gamma'=0.35$ safety gate~\cite{behrens2021painlimits}. \\
Physics-aware & C2 Joint-limit margin & $D_{limit}=\min(q-q_{\min},q_{\max}-q)$ for the \emph{simulated care recipient's joints} (human kinematic limits; not robot joint limits). \\
Task-level & C3 Success & Closed-loop task completion under geometric/contact gates, with failure attribution to perception, planning, execution, or safety truncation. \\
Physics-aware & C4 Motion activity & Active-motion fraction $\eta_{act}=T_{act}/T_{total}$ after human settle (pacing / temporal demand~\cite{hart1988nasatlx}). \\
Physics-aware & C5 Force stability & Within-run population SD of $\|F_c\|$ over the contact window (lower is more stable). Force--indentation coupling is reported in Section~\ref{sec:real_to_sim} rather than as a separate deformation score. \\
\bottomrule
\end{tabularx}
\end{table}

\subsection{Evaluation Protocol}
\label{sec:protocol}

Each evaluated method consumes the planner-visible snapshot interface of Section~\ref{sec:observation_interface}. State Machine derives its contact geometry from the initial RGB snapshot through conventional CV segmentation and scripted fitting of candidate contact keypoints and approach directions; no ground-truth coordinates or scorer-only state are provided to it. The protocol freezes the scorer--logger contract used for all trials in Section~\ref{sec:validation} (Table~\ref{tab:protocol_freeze}). Scorer-only streams never include future contact labels, and vision-facing methods never receive ground-truth joint angles or soft-anchor states.

\begin{table}[t]
\centering
\caption{Unified evaluation protocol freeze. All methods and selected tasks share this contract before metric aggregation.}
\label{tab:protocol_freeze}
\footnotesize
\begin{tabularx}{\columnwidth}{>{\raggedright\arraybackslash}p{0.28\columnwidth} >{\raggedright\arraybackslash}X}
\toprule
Item & Freeze rule \\
\midrule
Assets / embodiment & Fixed nursing-bed scene; same human and robot models across methods. \\
Randomization / seeds & Comfort-arm orientation in $[45^\circ,60^\circ]$; fixed seed set per task--method cell. \\
Trial budget & $20$ initializations $\times$ $7$ tasks $=$ $140$ runs per method. \\
Success / safety gates & Task-specific geometric and contact gates for C3; $\gamma'$-scaled safety force gates and human joint envelopes as safety context. \\
Logger schema & RGB-D, point clouds, masks, robot proprioception; scorer-only force / joint streams offline. \\
Replay & Metrics regenerated from logged trajectories under the same scorer binary. \\
NA rules & No valid contact $\Rightarrow$ C1/C2/C5 reported as NA and C3~$=$~$0$. Valid-data coverage is not C3. \\
\bottomrule
\end{tabularx}
\end{table}

\section{Experiments}
\label{sec:validation}

We first state how the benchmark is used, then ask whether the environment is physically valid, then whether it separates policies, then what task-level scores miss. Passive-joint response remains a required property of the care recipient, not an optional ablation switch.

\subsection{Benchmark Setup}
\label{sec:vla_evaluation}

Before comparing methods, we use scripted probes to check that the scorer--logger parameters respond as intended: increasing commanded push depth increases C1; violating joint envelopes decreases C2; injecting stop--go execution decreases C4; larger within-window force fluctuation increases C5. These probes validate the evaluation apparatus, not the human-model calibration of Section~\ref{sec:real_to_sim}. Policy runs below inherit the same seed and logging contract\label{sec:metric_reliability}\label{sec:repeatability}.

\paragraph{Evaluated methods.}
\label{sec:policy_evaluation}
We instantiate the benchmark with an LLM-augmented state machine (State Machine in tables) and VoxPoser on the full T1--T7 suite ($20$ initializations $\times$ $7$ tasks $=$ $140$ runs per method). The two subjects represent distinct policy paradigms---state-machine control versus spatial reasoning:
\begin{enumerate}
\item \textbf{LLM-augmented state machine (State Machine in tables)}: an LLM-augmented controller consistent with finite-state coordination used in assistive bathing systems such as I-SUPPORT~\cite{zlatintsi2020isupport}. The method uses the same leakage-free observation contract as the other methods. From an initial RGB snapshot, conventional CV segmentation estimates the target-region geometry; scripted routines then fit candidate contact keypoints and approach directions and instantiate a bounded set of predefined Cartesian motion primitives, from which the LLM selects according to the language instruction. No ground-truth simulator state or scorer-only signal is used. The method is vision-based at initialization but does not update target geometry during execution, making it a structured, snapshot-conditioned reference rather than an end-to-end learned vision policy.
\item \textbf{VoxPoser~\cite{huang2023voxposer}}: zero-shot 3D value-map trajectories from LLMs/VLMs, using the vision interface of Section~\ref{sec:observation_interface}.
\end{enumerate}
Separately, we evaluate the publicly released $\pi_{0.5}$ DROID checkpoint~\cite{black2025pi05} pretrained for Franka Panda generalization (\code{pi05\_droid}), applied zero-shot to this contact-rich assistive-care setting without additional teleoperation or fine-tuning. This run is a zero-shot transfer stress test, not a third peer paradigm.

\begin{figure}[!t]
\centering
\includegraphics[width=\columnwidth]{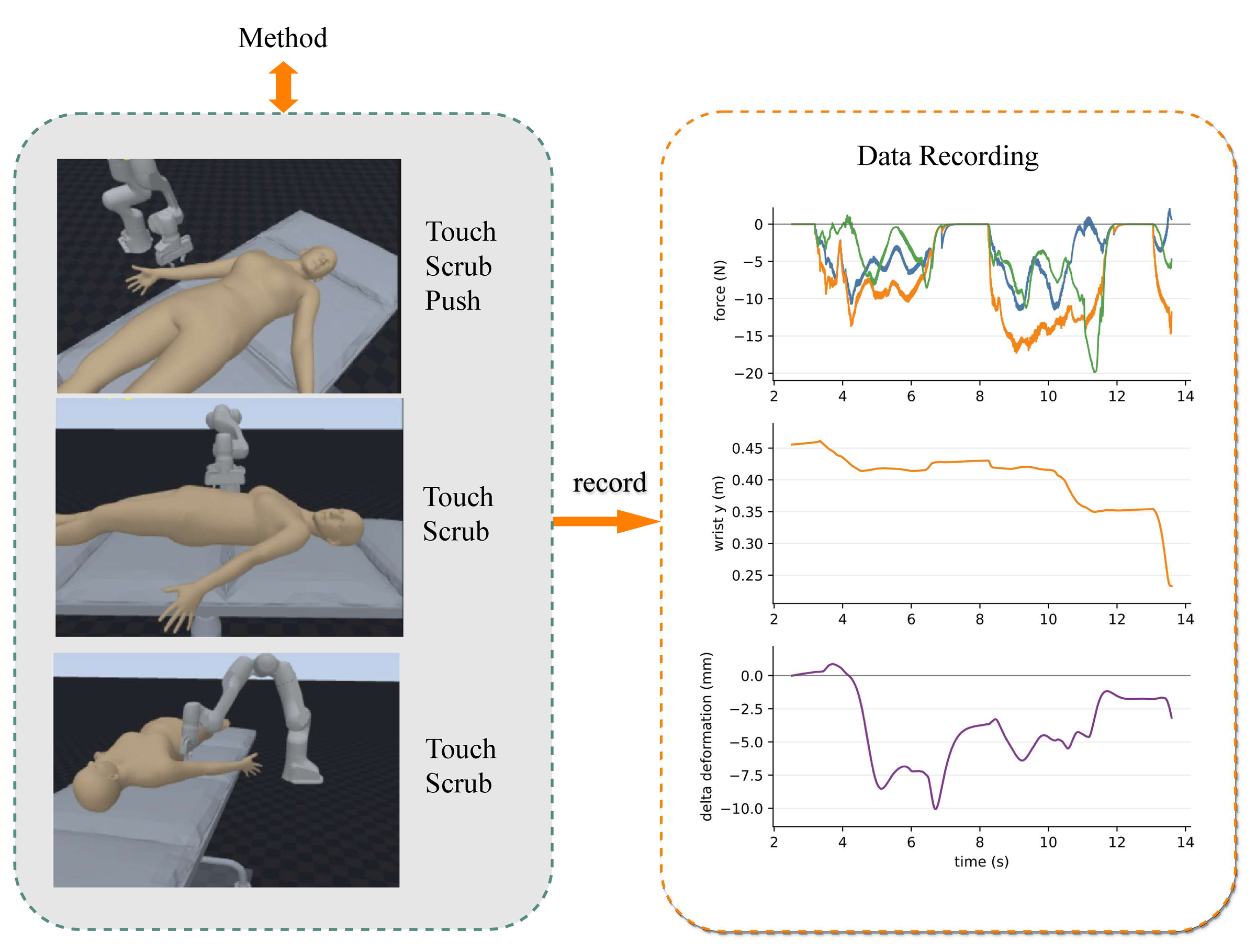}
\caption{Recorded language-conditioned interaction used by the scorer--logger. Left: Touch / Scrub / Push scenes on the forearm, upper arm, and abdomen. Right: logged contact force, wrist displacement, and local deformation. These traces illustrate the logging contract; deformation is not sold as a separate metric.}
\label{fig:exp_demo}
\end{figure}

Figure~\ref{fig:exp_demo} shows the recorded T1--T7 contact scenes and the force, wrist, and local-deformation streams written by the logger. Metrics follow Section~\ref{sec:metrics}. In later figures, \emph{valid ratio} is the fraction of runs that produced scoreable contact or trajectory data; it is not C3. Binary rates use Wilson $95\%$ confidence intervals; continuous metrics are reported as medians.

\subsection{Physical Validity of the Benchmark}
\label{sec:real_to_sim}

Real contact trials use a Franka Panda with a gripper under Cartesian impedance control. The contact partner is a medical-care manikin with a compliant surface and movable joints. Each trial records robot pose, contact force, and the displacement needed to reconstruct a force--indentation curve. We do \emph{not} instrument physical recovery after unload, so unload recovery and energy dissipation are excluded from the real-to-sim claim.

\begin{figure}[!t]
\centering
\includegraphics[width=\columnwidth]{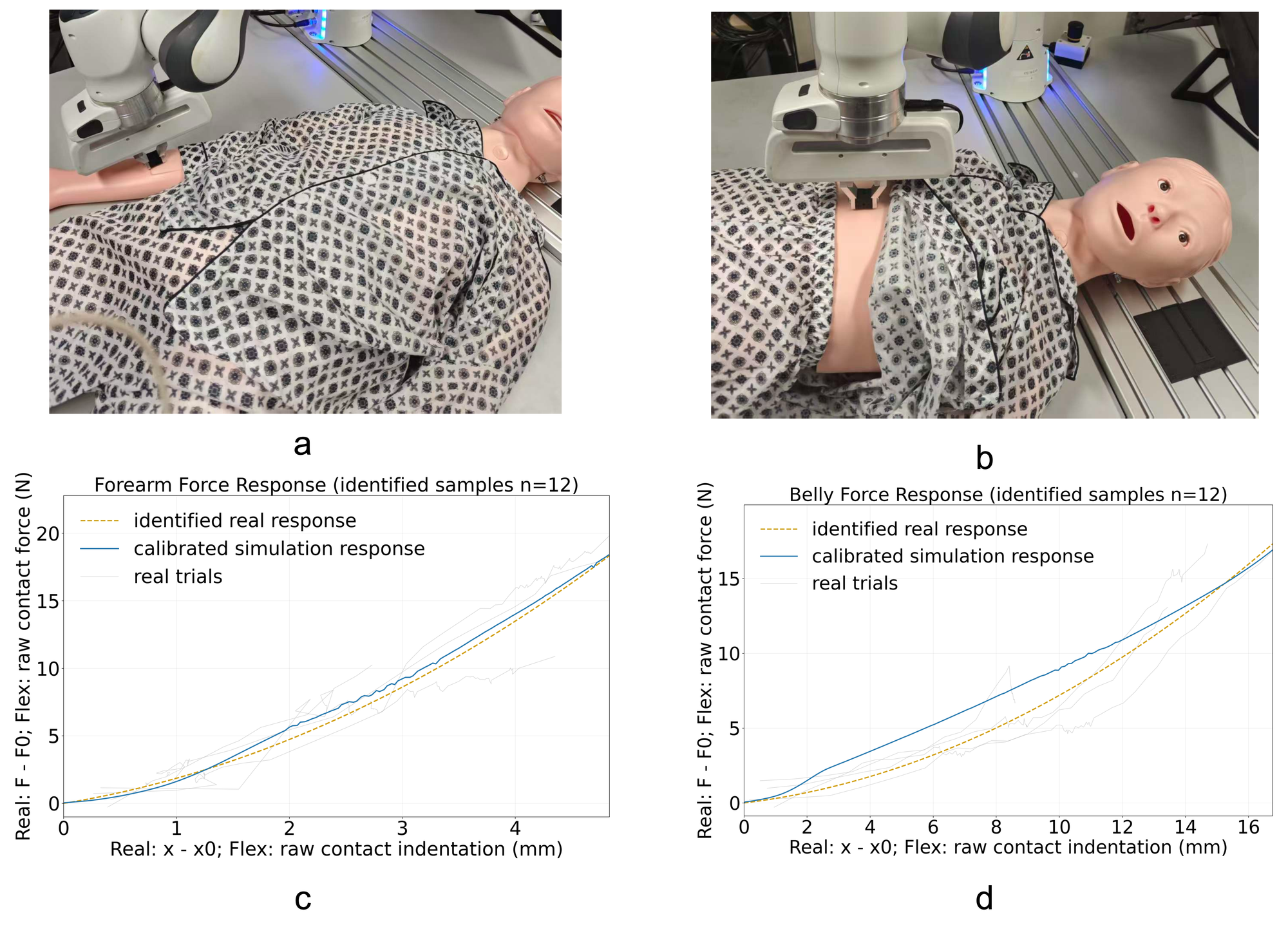}
\caption{Real-to-sim force--indentation calibration after impedance push on a medical-care manikin. (a)~Forearm push. (b)~Abdomen push. (c)~Forearm force--indentation. (d)~Abdomen force--indentation. Gray: measured trials. Gold dashed: identified response. Blue: calibrated simulation.}
\label{fig:r2s_fd}
\end{figure}

Matched pose schedules, contact locations, and directions are replayed in simulation. Figure~\ref{fig:r2s_fd} shows the forearm and abdomen pushes together with the overlaid force--indentation curves: raw measured trials (gray), the identified response (gold dashed), and the calibrated simulation (blue). Agreement on these curves is the evidence that Flex parameters are calibrated against measured contact responses. The same plots make the force--deformation coupling explicit, which is why deformation is not sold as a separate metric in Table~\ref{tab:metric_defs}. The upper-arm simulation reuses the forearm calibration parameters and is not independently calibrated against a real upper-arm response; manikin upper-arm compliance may differ from human tissue, so upper-arm C1/C2/C5 results are benchmark-level comparative scores rather than independent region-specific biomechanical validation.

\subsection{Benchmark Evaluation}
\label{sec:vla_results}

Table~\ref{tab:wilson_rates} organizes the method-level binary results as a progression from evaluator success, through correct-region grounding, to safety-gated deployment success. State Machine attains the highest task-level completion ($72.9\%$ $[65.0, 79.5]$). That result shows the scorer can recognize successful execution for a structured method in which language selects among vision-derived, script-generated Cartesian candidates. The same method carries a heavy force tail (median peak $19.48$\,N) and the second-worst C5 (Table~\ref{tab:policy_results}), so ``finished'' and ``stable contact'' remain distinct. The comparison characterizes distinct failure modes rather than establishing an architecture-neutral policy leaderboard.

\begin{table}[!t]
\centering
\caption{Method-level binary rates on T1--T7 ($140$ runs/method) with Wilson $95\%$ CIs. Safety-gated deployment success requires original evaluator success, correct-region contact, and peak force within the $\gamma'=0.35$ gate. Task-level intervals are reported in the supplementary material. State Machine uses an initial RGB snapshot and performs no execution-time visual replanning.}
\label{tab:wilson_rates}

\footnotesize
\setlength{\tabcolsep}{3pt}
\renewcommand{\arraystretch}{1.15}

\begin{tabularx}{\columnwidth}{
>{\raggedright\arraybackslash}p{0.20\columnwidth}
*{3}{>{\centering\arraybackslash}X}}
\toprule
Method &
Task success $\uparrow$ &
Region-correct $\uparrow$ &
Safety-gated $\uparrow$ \\
\midrule

State Machine$^{\dagger}$ &
\cellcolor{cellgreen}\makecell{72.9\% \\ \scriptsize(65.0, 79.5)} &
\cellcolor{cellgreen}\makecell{72.9\% \\ \scriptsize(65.0, 79.5)} &
\cellcolor{cellgreen}\makecell{56.4\% \\ \scriptsize(48.2, 64.4)} \\

VoxPoser &
\cellcolor{cellyellow}\makecell{27.9\% \\ \scriptsize(21.1, 35.8)} &
\cellcolor{cellyellow}\makecell{27.9\% \\ \scriptsize(21.1, 35.8)} &
\cellcolor{cellyellow}\makecell{27.9\% \\ \scriptsize(21.1, 35.8)} \\

$\pi_{0.5}$$^{\ddagger}$ &
\cellcolor{cellred}\makecell{0.7\% \\ \scriptsize(0.1, 3.9)} &
\cellcolor{cellred}\makecell{0.0\% \\ \scriptsize(0.0, 2.7)} &
\cellcolor{cellred}\makecell{0.0\% \\ \scriptsize(0.0, 2.7)} \\

\bottomrule
\end{tabularx}

\vspace{2pt}
\raggedright
\scriptsize
Cell colors indicate numerical ranking within each column:
\textcolor{cellgreen}{\rule{0.8em}{0.8em}} best,
\textcolor{cellyellow}{\rule{0.8em}{0.8em}} second,
\textcolor{cellred}{\rule{0.8em}{0.8em}} worst.
$\uparrow$/$\downarrow$ denotes higher/lower is better.
$^{\dagger}$State Machine uses an initial RGB snapshot to generate scripted Cartesian candidates and performs no execution-time visual replanning.
$^{\ddagger}$All recorded $\pi_{0.5}$ contacts are off-target.
\end{table}

\noindent\textit{Gate sensitivity.} We recompute safety-gated success under a continuous sweep of $\gamma'$ over $[0.20,1.00]$. VoxPoser's and $\pi_{0.5}$'s gated success remain unchanged throughout this range, at $27.9\%$ and $0.0\%$, respectively. State Machine's gated success increases monotonically with $\gamma'$---for example, $35.0\%$ at $\gamma'=0.20$, $56.4\%$ at $\gamma'=0.35$, $71.4\%$ at $\gamma'=0.70$, and $72.1\%$ at $\gamma'=1.00$---but remains below its $72.9\%$ nominal success and above VoxPoser's gated success throughout the sweep. We report $\gamma'=0.35$ as a representative conservative point within this range rather than a precisely calibrated value; the nominal--gated dissociation and method ranking are preserved across the full sweep.


C4 is an operational stop--go continuity proxy: a higher active-motion fraction indicates fewer execution interruptions. Robot-motion rhythm and predictability are workload-relevant in HRI~\cite{carissoli2024mentalworkload}; in passive physical contact, stop--go interruptions may therefore be perceived as unpredictable and plausibly increase recipient anxiety. This experiential link is not measured here: C4 quantifies motion continuity only, and the hypothesis remains for future work.
\begin{table}[!t]
\centering
\caption{Pooled T1--T7 physics-aware scores under the 140-run/method protocol.
Each statistic uses available scoreable runs, so valid $n$ varies by metric and method.}
\label{tab:policy_results}

\footnotesize
\setlength{\tabcolsep}{3pt}
\renewcommand{\arraystretch}{1.15}

\begin{tabularx}{\columnwidth}{
>{\raggedright\arraybackslash}p{0.20\columnwidth}
*{4}{>{\centering\arraybackslash}X}}
\toprule
Method &
C1 $\downarrow$ &
C2 $\uparrow$ &
C4 $\uparrow$ &
C5 $\downarrow$ \\
\midrule

State Machine &
\cellcolor{cellred}$19.48$ &
\cellcolor{cellgreen}$-3.61{\times}10^{-4}$ &
\cellcolor{cellred}$0.788$ &
\cellcolor{cellyellow}$3.52$ \\

VoxPoser &
\cellcolor{cellgreen}$1.46$ &
\cellcolor{cellyellow}$-4.07{\times}10^{-4}$ &
\cellcolor{cellyellow}$0.883$ &
\cellcolor{cellgreen}$0.21$ \\

$\pi_{0.5}$$^{\dagger}$ &
\cellcolor{cellyellow}$15.39$ &
\cellcolor{cellred}$-1.37{\times}10^{-3}$ &
\cellcolor{cellgreen}$0.999$ &
\cellcolor{cellred}$4.42$ \\

\bottomrule
\end{tabularx}

\vspace{2pt}
\raggedright
\scriptsize
C1: median peak contact force (N);
C2: median signed human joint-limit margin on Push (rad);
C4: median active-motion fraction (higher indicates less stop--go execution);
C5: median within-run force SD (N).
$\uparrow$/$\downarrow$ denotes higher/lower is better.
Cell colors indicate numerical ranking within each directional metric:
\textcolor{cellgreen}{\rule{0.8em}{0.8em}} best,
\textcolor{cellyellow}{\rule{0.8em}{0.8em}} second,
\textcolor{cellred}{\rule{0.8em}{0.8em}} worst.
$^{\dagger}$For $\pi_{0.5}$, C1/C5 statistics come entirely from off-target-region contact.
\end{table}

\begin{table}[!t]
\centering
\caption{Pre-specified State Machine versus VoxPoser contrasts under the $140$-run/method protocol. Binary tests use all trials; continuous tests use available scoreable runs. These are not a full pairwise screen; $\pi_{0.5}$ is omitted.}
\label{tab:cross_tests}
\footnotesize
\begin{tabularx}{\columnwidth}{>{\raggedright\arraybackslash}X >{\raggedright\arraybackslash}p{0.28\columnwidth} >{\centering\arraybackslash}p{0.22\columnwidth}}
\toprule
Outcome & Test & $p$ \\
\midrule
Correct-region success & Fisher exact & $4.74\times10^{-14}$ \\
Safety-gated success & Fisher exact & $1.99\times10^{-6}$ \\
Peak contact force & Mann--Whitney $U$ & $5.72\times10^{-40}$ \\
Within-run force SD & Mann--Whitney $U$ & $4.75\times10^{-40}$ \\
\bottomrule
\end{tabularx}
\end{table}

\begin{figure*}[!t]
\centering
\includegraphics[width=\textwidth]{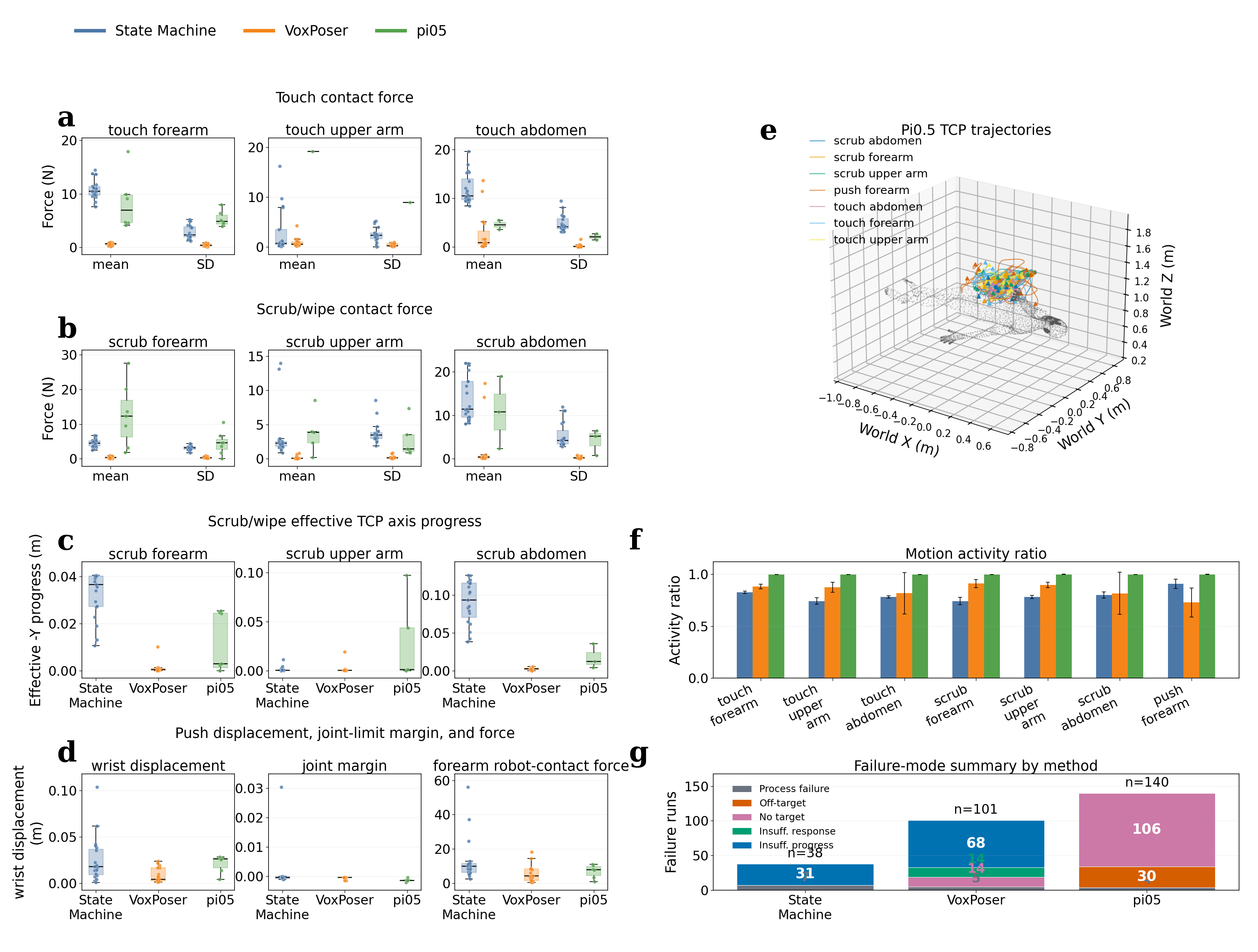}
\caption{Language-conditioned evaluation on T1--T7. (a)~Touch contact force (run mean and within-run SD). (b)~Scrub contact force. (c)~Scrub effective stroke. (d)~Push displacement, human joint-limit margin, and forearm force. (e)~$\pi_{0.5}$ TCP trajectories. (f)~Motion-activity ratio (C4). (g)~Primary failure mode by method. Valid ratio is scoreable-data coverage, not closed-loop success. Force and TCP panels for $\pi_{0.5}$ show off-target-region contact. In (g) the denominator is failure runs after excluding correct-region successes ($38$ / $101$ / $140$ for State Machine / VoxPoser / $\pi_{0.5}$); the single original-evaluator success of $\pi_{0.5}$ is off-target and remains in the failure analysis.}
\label{fig:exp_method_outcomes}
\end{figure*}

VoxPoser produces the lightest and most stable conditional contact (C1 $1.46$\,N, C5 $0.21$\,N) and a high motion-activity score, yet converts this into C3 only $27.9\%$ of the time, concentrated on Touch. State Machine and VoxPoser differ in correct-region success and contact-force statistics (Table~\ref{tab:cross_tests}). Low force here is therefore not a contact skill: the motion is often too weak to complete Push or Scrub (Fig.~\ref{fig:exp_method_outcomes}(a)--(d)).

The near-zero success rate of $\pi_{0.5}$ ($0.7\%$) is consistent with substantial sensitivity to zero-shot transfer from the DROID / Franka pretraining distribution into this setting. Its scoreable contact set is small ($30$ force runs, $6$ Push-margin runs), and every recorded contact is on a \emph{non-target} body region. The C1/C5 numbers for $\pi_{0.5}$ must not be read as target-region contact forces. Figure~\ref{fig:exp_method_outcomes}(e) shows TCP trajectories that fail to lock onto the instructed region.

\subsection{What Conventional Evaluation Misses}

For failure-mode analysis, we condition on trials that fail the region-correctness criterion; percentages therefore use these failed trials rather than all trials as the denominator ($38$ / $101$ / $140$ for State Machine / VoxPoser / $\pi_{0.5}$).

Region-wise contact (Table~\ref{tab:contact_anatomy}) separates reaching the instructed site from finishing the action. State Machine and VoxPoser both establish target-region contact on $95.0\%$ of forearm runs, yet VoxPoser's original-evaluator success on that same forearm task set is only $33.3\%$: contact is not completion. $\pi_{0.5}$ records $0$ target-region contacts in all three sites. Valid ratio in Fig.~\ref{fig:exp_method_outcomes} remains scoreable-data coverage, not C3.

\begin{table}[!t]
\centering
\caption{Region-wise contact outcomes on T1--T7. Rankings are evaluated
within each body region.}
\label{tab:contact_anatomy}

\footnotesize
\setlength{\tabcolsep}{3pt}
\renewcommand{\arraystretch}{1.12}

\begin{tabularx}{\columnwidth}{
>{\raggedright\arraybackslash}p{0.16\columnwidth}
>{\raggedright\arraybackslash}p{0.16\columnwidth}
*{4}{>{\centering\arraybackslash}X}}
\toprule
Region &
Method &
Target $\uparrow$ &
Off-tgt $\downarrow$ &
No-contact $\downarrow$ &
C3 $\uparrow$ \\
\midrule

Forearm &
State Machine &
\cellcolor{cellgreen}$57/60$ &
\cellcolor{cellgreen}$0$ &
\cellcolor{cellgreen}$0$ &
\cellcolor{cellgreen}$54/60$ \\

Forearm &
VoxPoser &
\cellcolor{cellgreen}$57/60$ &
\cellcolor{cellgreen}$0$ &
\cellcolor{cellyellow}$2/60$ &
\cellcolor{cellyellow}$20/60$ \\

Forearm &
$\pi_{0.5}$ &
\cellcolor{cellred}$0$ &
\cellcolor{cellred}$19/60$ &
\cellcolor{cellred}$40/60$ &
\cellcolor{cellred}$1/60$ \\

\cmidrule(lr){1-6}

Upper arm &
State Machine &
\cellcolor{cellgreen}$37/40$ &
\cellcolor{cellgreen}$0$ &
\cellcolor{cellgreen}$0$ &
\cellcolor{cellgreen}$25/40$ \\

Upper arm &
VoxPoser &
\cellcolor{cellyellow}$34/40$ &
\cellcolor{cellgreen}$0$ &
\cellcolor{cellyellow}$3/40$ &
\cellcolor{cellyellow}$11/40$ \\

Upper arm &
$\pi_{0.5}$ &
\cellcolor{cellred}$0$ &
\cellcolor{cellred}$6/40$ &
\cellcolor{cellred}$32/40$ &
\cellcolor{cellred}$0$ \\

\cmidrule(lr){1-6}

Abdomen &
State Machine &
\cellcolor{cellgreen}$39/40$ &
\cellcolor{cellgreen}$0$ &
\cellcolor{cellgreen}$0$ &
\cellcolor{cellgreen}$23/40$ \\

Abdomen &
VoxPoser &
\cellcolor{cellyellow}$30/40$ &
\cellcolor{cellgreen}$0$ &
\cellcolor{cellyellow}$9/40$ &
\cellcolor{cellyellow}$8/40$ \\

Abdomen &
$\pi_{0.5}$ &
\cellcolor{cellred}$0$ &
\cellcolor{cellred}$5/40$ &
\cellcolor{cellred}$34/40$ &
\cellcolor{cellred}$0$ \\

\bottomrule
\end{tabularx}

\vspace{2pt}
\raggedright
\scriptsize
Target: target-region contact;
Off-tgt: off-target contact;
No-contact: completed without contact;
C3: original evaluator success.
$\uparrow$/$\downarrow$ denotes higher/lower is better.
Cell colors indicate numerical ranking within each body region:
\textcolor{cellgreen}{\rule{0.8em}{0.8em}} best,
\textcolor{cellyellow}{\rule{0.8em}{0.8em}} second,
\textcolor{cellred}{\rule{0.8em}{0.8em}} worst; ties share the same color.
Denominators are 60 runs for forearm and 40 runs each for upper arm and abdomen.
Process failures are omitted from these columns:
State Machine $3/3/1$, VoxPoser $1/3/1$, and $\pi_{0.5}$ $1/2/1$
for forearm/upper arm/abdomen, respectively.
All established $\pi_{0.5}$ contacts are off-target.
\end{table}

The main diagnostic is VoxPoser. Region-wise scores show a systematic gap between reaching a site and sustaining contact: on Touch, target-region contact is comparable to State Machine, yet Scrub and Push completion collapse, and the dominant failure is insufficient progress ($68/101$) rather than localization error. High motion activity (C4 $0.883$) does not convert into task progress, consistent with persistent but poorly directed motion around a safety-oriented objective; C4 and C3 together show that a single score cannot decide effectiveness. Value-map methods that treat a safety-proxy metric as the sole objective can therefore reduce contact to a safe but ineffective regime. A concrete implication is that the value map's role may need to narrow once contact is established: rather than continuing to shape the trajectory toward a low-cost configuration, a later phase can be driven by an explicit progress objective---for example a target sliding distance of the kind already reported as effective stroke for Scrub (Fig.~\ref{fig:exp_method_outcomes}(c)), or a target indentation of the kind already reported as wrist displacement for Push (Fig.~\ref{fig:exp_method_outcomes}(d))---so that sustained contact is driven by a progress signal rather than a safety proxy. Reaching and sustained-contact phases may therefore warrant distinct control regimes rather than a single value-map objective throughout.

State Machine supplies a shorter supporting point: even with a structured candidate generator that deliberately narrows the action space, it still carries a force tail and drops from $102/140$ to $79/140$ under the safety gate ($36/140$ exceedances, of which only a subset were originally successful and region-correct). The benchmark therefore still surfaces physical problems in a constrained reference with strong geometric scaffolding.

The $\pi_{0.5}$ outcome remains a short transfer probe: $1/140$ original success, $0/140$ correct-region success, and $0/140$ gated success, consistent with sensitivity to zero-shot transfer from the DROID / Franka pretraining distribution. Its failures are almost entirely no target contact ($106/140$) plus off-target contact ($30/140$). Safe interaction is therefore not the same as effective interaction. More broadly, task completion does not imply physically valid contact. The benchmark can serve as a pre-deployment physical/safety screen for contact-rich care policies.

\section{Conclusion}
\label{sec:conclusion}

This study uses $20$ trials per task and is intended for benchmark characterization rather than population-level statistical estimation. The contact partner is a medical-care manikin, not human tissue; the upper-arm simulation reuses the forearm calibration; $K_c$/$D_c$ sensitivity is not systematically studied; and the evaluated set is limited to a LLM-augmented state-machine reference, a spatial-reasoning planner, and one zero-shot Franka-DROID checkpoint. State Machine instead builds a bounded set of scripted Cartesian candidates from an initial RGB snapshot and lets the LLM select among them, without execution-time visual replanning or privileged simulator state. The instantiation is a single robotic-bathing scenario.

Within that scope, conventional task-level evaluation can hide contact failures. The physics-consistent benchmark makes those dissociations measurable: State Machine can finish with an unstable force tail, VoxPoser can remain light without completing sustained contact, and $\pi_{0.5}$ can move continuously while remaining off-target. Task completion does not imply physically valid contact. The benchmark can serve as a pre-deployment physical/safety screen for contact-rich care policies.

\FloatBarrier
\bibliographystyle{IEEEtran}
\bibliography{references_ieee}

\end{document}